\documentclass[11pt,a4paper]{article}
\usepackage[hyperref]{acl} 
\usepackage{times}
\usepackage{tabularray}
\usepackage{comment}
\usepackage{float}
\UseTblrLibrary{booktabs}
\usepackage{latexsym}
\usepackage{graphicx}
\usepackage{amssymb}
\usepackage{pifont}

\usepackage{booktabs}
\usepackage{array}

\usepackage{booktabs}
\usepackage{multirow}
\usepackage{gb4e}  
\usepackage{enumitem}
\noautomath

\usepackage{tipa}

\usepackage{microtype}
\usepackage{soul}
\usepackage[whole]{bxcjkjatype}

\title{Transformed Protoform Reconstruction}

\author{Young Min Kim\Thanks{ Equal contribution} \and Kalvin Chang\footnotemark[1] \and Chenxuan Cui \and David Mortensen \\
        Language Technologies Institute, Carnegie Mellon University \\
        \texttt{ \{youngmik, kalvinc, cxcui, dmortens\}@cs.cmu.edu }}

\date{}

\begin{document}
\maketitle
\begin{abstract}

Protoform reconstruction is the task of inferring how morphemes or words sounded in ancestral languages of a set of daughter languages. \citet{meloni} achieved the state-of-the-art on Latin protoform reconstruction with an RNN-based encoder-decoder with attention model. 
We update their model with the state-of-the-art seq2seq model---the Transformer. Our model outperforms their model on a suite of different metrics on two different datasets: \citeauthor{meloni}'s Romance data of 8,000+ cognates (spanning 5 languages) and a Chinese dataset \citep{Hou2004} of 800+ cognates (spanning 39 varieties). 
We also probe our model for potential phylogenetic signal contained in the model. 
Our code is publicly available \footnote{\url{https://github.com/cmu-llab/acl-2023}}.


\end{abstract}

\section{Introduction}

    Languages change over time and sometimes diverge into multiple daughter languages. The common ancestor of a set of genetically related languages is their proto-language. While there are proto-languages such as Latin that are attested, they are the exception\footnote{In fact, the proto-language from which Romance languages like Spanish and Italian are descended is not identical to Classical Latin but is, rather, a closely related and sparsely attested language sometimes called Proto-Romance or Vulgar Latin.}.
    Reconstructed words and morphemes in proto-languages are called protoforms. The task of reconstructing unattested proto-languages is called protoform reconstruction. 

    Historical linguists reconstruct proto-languages by identifying systematic sound changes that can be inferred from correspondences between attested daughter languages (see Table~\ref{tab:sound-coresp-west-germ}). They compare the sounds between a set of cognates, or words with a common ancestor, to develop hypotheses about the types and chronologies of sound changes.

    \begin{table}[htb]
      \centering
      \SetTblrInner{rowsep=0pt}
      \begin{tblr}{lr@{}lr@{}lr@{}lr}
        \toprule
        & & `tooth' & & `two' & & `ten' & \\
        \midrule
        English & & \textbf{t}ooth & & \textbf{t}wo & & \textbf{t}en & \textbf{t}\\
        Dutch & & \textbf{t}and & & \textbf{t}wee & & \textbf{t}ien & \textbf{t}\\
        German & & \textbf{Z}ahn & & \textbf{z}wei & & \textbf{z}ehn & \textbf{z}\\
        {PWG} & * & \textbf{t}an\textipa{þ} & * & \textbf{t}wai- & * & \textbf{t}ehun & \textbf{*t}\\
        \bottomrule
      \end{tblr}
      \caption{Sound correspondences in West Germanic Languages and Proto-West-Germanic (PWG).}
      \label{tab:sound-coresp-west-germ}
    \end{table}

    This task is inherently data-constrained, especially for under-documented languages. Such data scarcity makes it a particularly difficult task for contemporary neural network architectures such as the Transformer \cite{attention-is-all-you-need}, which are data hungry.

    The contributions of this paper are as follows:
    \begin{itemize}
    \item Application of the Transformer architecture to the protoform reconstruction task, achieving state of the art performance, contrary to expectation. 
    \item Expansion of prior digital versions of \citet{Hou2004}'s Chinese dataset to include a total of 804 cognate sets across 39 modern varieties and Middle Chinese. 
    \end{itemize}
    


\section{Related Work}

    Applying machine learning to protoform reconstruction is not new. \citet{bouchard-cote-2013} learn an unsupervised protoform reconstruction model for the large Oceanic language family using Monte Carlo Expectation Maximization \cite{em,mc-inference}, supervising the model with a gold phylogeny and using a probabilistic, generative model of sound change. \citet{he2022neural} modernize an earlier version of \citet{bouchard-cote-2013}'s model with RNNs for a 4 language subset of Romance, but they rely on a bigram language model of Latin, making their model technically not unsupervised.

    \citet{list-svm} apply an SVM classifier to supervised reconstruction by treating sound correspondences as training examples. Note that there were no word boundaries in the input matrix; that is, all sound correspondences across the training set are flattened into one matrix. Furthermore, each language has an independent phonemic inventory. To learn contextual information, the authors experiment with adding features encoding the position of phonemes, among others.
    
    \citet{ciobanu-dinu-2018-ab} learn a conditional random field \citep{crf} using n-gram features for supervised reconstruction and ensemble 5 daughter-to-protoform models. They use a dataset of 3,218 complete cognate sets spanning Latin (the proto-language) and 5 Romance languages: Romanian, French, Italian, Spanish, Portuguese. 

    \citet{meloni} employ a GRU-based seq2seq approach \citep{encoder-decoder} to Latin protoform reconstruction and achieve state-of-the-art character edit distances. They extend \citet{ciobanu-dataset}'s Romance data using data from Wiktionary---for a total of 8,799 cognate sets across 5 Romance languages plus Latin---in both orthographic and phonetic (IPA) representations. In their model, all entries comprising the cognate set are concatenated together in a fixed order to form a training example. \citet{wikihan} applied \citet{meloni}'s architecture to the reconstruction of Middle Chinese on a dataset of 5000+ cognate sets spanning 8 languages they compiled from Wiktionary.\footnote{The original dataset contains 21,000 cognate sets, but only 5000+ had at least 3 daughter entries and were used as input to the model.}

    \citet{fourrier-thesis} compares statistical machine translation, RNN, and Transformer architectures for protoform reconstruction, but they evaluate their results using BLEU scores \citep{bleu} instead of edit distance. They find that their Transformer model did not outperform the RNN models on protoform reconstruction. In addition, their multilingual NMT (neural machine translation) model predicts many languages instead of one target language and is trained on bilingual pairs for protoform reconstruction (e.g. Italian-Latin and Spanish-Latin), unlike comparative reconstruction. In contrast, we encode the entire cognate set consisting of multiple daughter languages (5 for the Romance dataset; 39 for Chinese) and predict the corresponding protoform.

\section{Datasets}

    We train and test our model on Romance and Sinitic (Chinese) language datasets. For Romance languages, we use \citet{meloni}'s dataset which consists of 8,799 cognate sets of Romanian, French, Italian, Spanish, Portuguese words and the corresponding Latin form (approximately, a protoform). There are two versions of this dataset: phonetic and orthographic. The phonetic dataset (Rom-phon) represents words with IPA symbols whereas the orthographic dataset (Rom-orth) represents words in the orthographic form of each language. We preserved all diacritics, except for vowel length. This dataset is an extension of \citet{ciobanu-dataset}'s original dataset of 3,218 cognate sets, which is not publicly available. Refer to \autoref{table:datasets} for more information. 

\subsection{Expanding digital versions of \citet{Hou2004}}

    For Sinitic languages, we created a dataset of Middle Chinese and its modern daughter languages. Middle Chinese is an unattested language, and we thus have to rely on \citet{baxter}'s reconstructions of forms corresponding to 4,967 Chinese characters. We scraped Wiktionary to obtain \citet{Hou2004}'s phonetic representations of their modern reflexes.\footnote{\url{https://en.wiktionary.org/wiki/Module:zh/data/dial-pron/documentation} originally had 1,023 characters, but only 804 had reconstructions from \citet{baxter}.} The resulting dataset contains 804 cognate sets of 39 modern Sinitic languages and the corresponding reconstructed Middle Chinese word. \citet{hou-cldf}'s version previously had 894 cognate sets across 15 varieties.

\section{Model}
    We propose a Transformer-based encoder-decoder architecture \citep{attention-is-all-you-need} because such models have produced state-of-the-art results on many sequence processing tasks.  
    Transformers are by reputation data hungry, though, which poses a challenge to our problem setting, where the number of available training examples is often very small.
    \begin{figure}[H]
      \centering
      \includegraphics[width=0.8\linewidth]{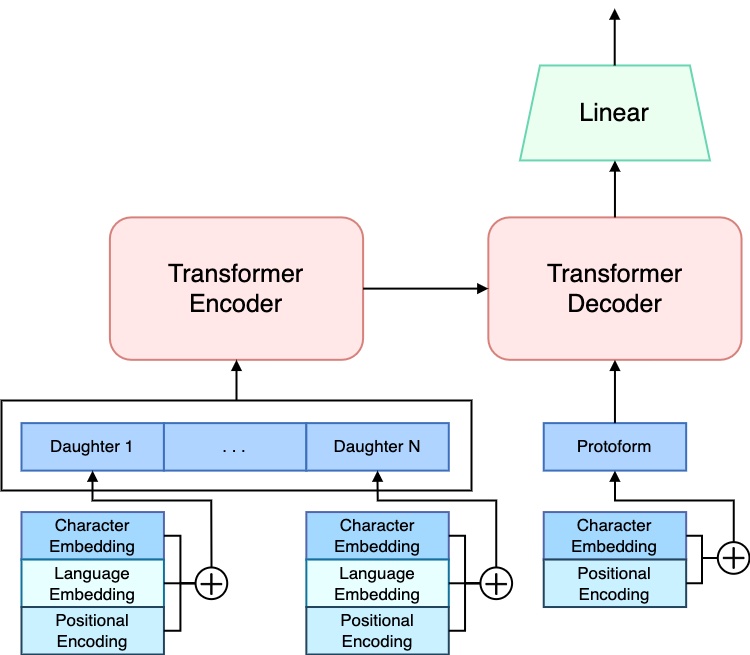}
      \caption{Diagram of our encoder-decoder architecture. Additive positional encoding and language embedding are applied to each daughter sequence before all daughter sequences are concatenated into a single sequence.
      }
      \label{fig:transformer}
    \end{figure}

    We modify the standard encoder-decoder architecture to accommodate the structure of our datasets, where multiple daughter sequences correspond to a single protoform sequence. Like \citet{meloni}, the daughter sequences are concatenated into a single sequence before being fed into the encoder. Because we only care about the relative position between tokens within each daughter sequence but not across daughter sequences, positional encoding is applied to each individual daughter sequence before concatenation. Along with positional encoding, an additive language embedding is applied to the token embeddings to differentiate between input tokens of different daughter languages. 



\begin{table*}
  \SetTblrInner{rowsep=0pt}
\begin{tblr}{lXlll}
\toprule
\textbf{Language Family} & \textbf{Source} & \textbf{\# varieties} & \textbf{Cognate sets} & \textbf{Proto-language} \\
\midrule
Rom-phon & \citet{ciobanu-dataset}, & 5 & 8,799 & Latin\\
& \citet{meloni} & & & \\
Rom-orth & \citet{ciobanu-dataset}, & 5 & 8,799 & Latin\\
& \citet{meloni} & & & \\
Sinitic (Chinese) & \citet{Hou2004} & 39 & 804 & Middle Chinese\\
\bottomrule
\end{tblr}
\caption{\label{table:datasets}
Statistics on both datasets used in our experiments. \# varieties refers to the number of daughter varieties.}
\end{table*}

\begin{table*}[htb]
  \centering
  \SetTblrInner{rowsep=1pt}
  \begin{tblr}{
      colspec={lXllllll}
    }
    \toprule
\textbf{Dataset} & \textbf{Model} & \textbf{PED $\downarrow$} & \textbf{NPED $\downarrow$} & \textbf{Acc \% $\uparrow$} & \textbf{FER $\downarrow$} & \textbf{BCFS $\uparrow$}  \\
    \midrule
    Sinitic & Random daughter \cite{wikihan} & 3.7702 & 0.8405 & 0\% & 0.2893 & 0.2748 \\
    & Majority constituent \cite{wikihan} & 3.5031 & 0.7806 & 0\% & 0.2013 & 0.3695 \\
    & CorPaR \cite{list-svm} & 3.2795 & 0.7278 & 0\% & 0.3972 & 0.3332 \\
    & SVM + PosStr \cite{list-svm} & 1.6894 & 0.3692 & 15.52\% & 0.1669 & 0.5418 \\
    & RNN \citep{meloni} & 1.0720 & 0.2432 & 35.47\% & 0.0896 & 0.6747 \\
    & Transformer (present work) & \textbf{0.9814} & \textbf{0.2204} & \textbf{39.50\%} & \textbf{0.0857} & \textbf{0.6971} \\
    
    \midrule
    Rom-phon & Random daughter \cite{wikihan} & 6.1534 & 0.6914 & 0.06\% & 0.6264 & 0.4016 \\
    & CorPaR + PosIni \cite{list-svm} & 1.6847 & 0.1978 & 22.18\% & 0.0728 & 0.7403 \\
    & SVM + PosStrIni \cite{list-svm} & 1.5787 & 0.1861 & 24.69\% & 0.0713 & 0.7610   \\
    & RNN \citep{meloni} & 0.9670 & 0.1229 & 52.09\% & 0.0385 & 0.8293 \\
    & Transformer (present work) & \textbf{0.9027} & \textbf{0.1146} & \textbf{53.16\% } & \textbf{0.0378 } & \textbf{0.8421} \\
    
    \midrule
    Rom-orth & Random daughter \cite{wikihan} & 4.2567 & 0.4854 & 2.97\% & $-$ & 0.5147  \\
    & CorPaR + Ini \cite{list-svm} & 0.9531 & 0.1160 & 47.23\% & $-$ & 0.8400 \\
    & SVM + PosStr \cite{list-svm} & 0.8988 & 0.1105 & 50.43\% & $-$ & 0.8501 \\
    & RNN & 0.5958 & 0.0772 & 69.74 \% & $-$ & 0.8913 \\
    & Transformer (present work) & \textbf{0.5568 } & \textbf{0.0724 } & \textbf{71.15\% } & $-$ & \textbf{0.8994 } \\
    
    \bottomrule
  \end{tblr}
  \caption[Results]{Evaluation of models and baselines using various metrics, averaged across 10 runs (same hyperparameters, different seeds). Because Rom-orth is not in IPA, character edit distance is used instead of PED, and we cannot accurately calculate FER. See \autoref{sec:eval} for an explanation of each evaluation metric. See \autoref{tab:results-full} for the standard deviation values.
  }
  \label{tab:results}
\end{table*}

\begin{table*}[htb]
  \centering
  \SetTblrInner{rowsep=1pt}
  \begin{tblr}{
      colspec={lXXXXXXX}
    }
    \toprule
\textbf{Dataset} & \textbf{Model} & \textbf{PED $\downarrow$} & \textbf{NPED $\downarrow$} & \textbf{Acc \% $\uparrow$} & \textbf{FER $\downarrow$} & \textbf{BCFS $\uparrow$}  \\
    \midrule
    Sinitic & Random daughter & 3.7702 & 0.8405 & 0\% & 0.2893 & 0.2748 \\
    & Majority constituent & 3.5031 & 0.7806 & 0\% & 0.2013 & 0.3695 \\
    & CorPaR & 3.2795 & 0.7278 & 0\% & 0.3972 & 0.3332 \\
    & SVM +PosStr & 1.6894 & 0.3692 & 15.52\% & 0.1669 & 0.5418 \\
    & RNN & 1.0720 $\pm$ 0.0536 & 0.2432 $\pm$ 0.0121 & 35.47\% $\pm$ 1.40\% & 0.0896 $\pm$ 0.0042 & 0.6747 $\pm$ 0.0166 \\
    & Transformer (present work) & \textbf{0.9814 $\pm$ 0.0437} & \textbf{0.2204 $\pm$ 0.0093} & \textbf{39.50\% $\pm$ 3.02\%} & \textbf{0.0857 $\pm$ 0.0057} & \textbf{0.6971 $\pm$ 0.0102} \\
    \midrule
    Rom-phon & Random daughter & 6.1534 & 0.6914 & 0.06\% & 0.6264 & 0.4016 \\
    & CorPaR +PosIni & 1.6847 & 0.1978 & 22.18\% & 0.0728 & 0.7403 \\
    & SVM +PosStrIni & 1.5787 & 0.1861 & 24.69\% & 0.0713 & 0.7610   \\
    & RNN & 0.9670 $\pm$ 0.0163 & 0.1229 $\pm$ 0.0020 & 52.09\% $\pm$ 0.59\% & 0.0385 $\pm$ 0.0011 & 0.8293 $\pm$ 0.0024 \\
    & Transformer (present work) & \textbf{0.9027 $\pm$ 0.0194} & \textbf{0.1146 $\pm$ 0.0021} & \textbf{53.16\% $\pm$ 0.66\%} & \textbf{0.0378 $\pm$ 0.0011} & \textbf{0.8421 $\pm$ 0.0029} \\
    \midrule
    Rom-orth & Random daughter & 4.2567 & 0.4854 & 2.97\% & $-$ & 0.5147  \\
    & CorPaR +Ini & 0.9531 & 0.1160 & 47.23\% & $-$ & 0.8400 \\
    & SVM +PosStr & 0.8988 & 0.1105 & 50.43\% & $-$ & 0.8501 \\
    & RNN & 0.5958 $\pm$ 0.0083 & 0.0772 $\pm$ 0.0013 & 69.74 \% $\pm$ 0.23\% & $-$ & 0.8913 $\pm$ 0.0016 \\
    & Transformer (present work) & \textbf{0.5568 $\pm$ 0.0086} & \textbf{0.0724 $\pm$ 0.0013} & \textbf{71.15\% $\pm$ 0.38 \%} & $-$ & \textbf{0.8994 $\pm$ 0.0015} \\
    \bottomrule
  \end{tblr}
  \caption[Results]{Evaluation of models and baselines using various metrics, averaged across 10 runs (same hyperparameters, different seeds), with standard deviations. Because Rom-orth is not in IPA, character edit distance is used instead of PED, and we cannot accurately calculate FER. See \autoref{sec:eval} for an explanation of each evaluation metric. 
  }
  \label{tab:results-full}
\end{table*}

\section{Experiments}


\subsection{Baselines}
We compare our Transformer model to a variety of baselines.
For \citet{meloni}, we use \citet{wikihan}'s PyTorch re-implementation and reran a Bayesian hyperparameter search using WandB \citep{wandb} to ensure a more fair comparison (since our model is tuned with WandB as well). 
We also include the random daughter (randomly designate a daughter form as the protoform and assume no sound change) and the majority constituent baselines (predict the most common phoneme in each syllable constituent) from \citet{wikihan}. 
For the SVM and CoRPaR classifiers \citep{list-svm}, we experiment with different contextual features, such as Pos (position), Str (prosodic structure), and Ini (whether or not the phoneme appears word-initially or word-finally). 

We publish results on \citet{meloni}'s full set of 8,799 cognates but cannot redistribute this set due to \citet{ciobanu-dataset}'s restrictions. For reproducibility, we include results on \citet{meloni}'s public subset of 5,419 cognates in the Appendix (\autoref{tab:results-meloni-5000}), both of which include vowel length. Observe that these results are worse than those obtained on the full set, suggesting that the RNN and Transformer are dependent on a wealth of training data. 

\subsection{Preprocessing}
In all our datasets, we merge diacritics to their base segments to form a multi-character token. For instance, the sequence [t, \textsuperscript{h}] is concatenated to [t\textsuperscript{h}]. This ensures that phonemes are treated as one token. 
For Chinese, tone contours (a sequence of tones) are treated as one token. 
When multiple pronunciation variants are listed for a single Chinese character, we arbitrarily pick the first one. 

\begin{figure*}[h!]
    \centering
    \includegraphics[width=0.99\linewidth]{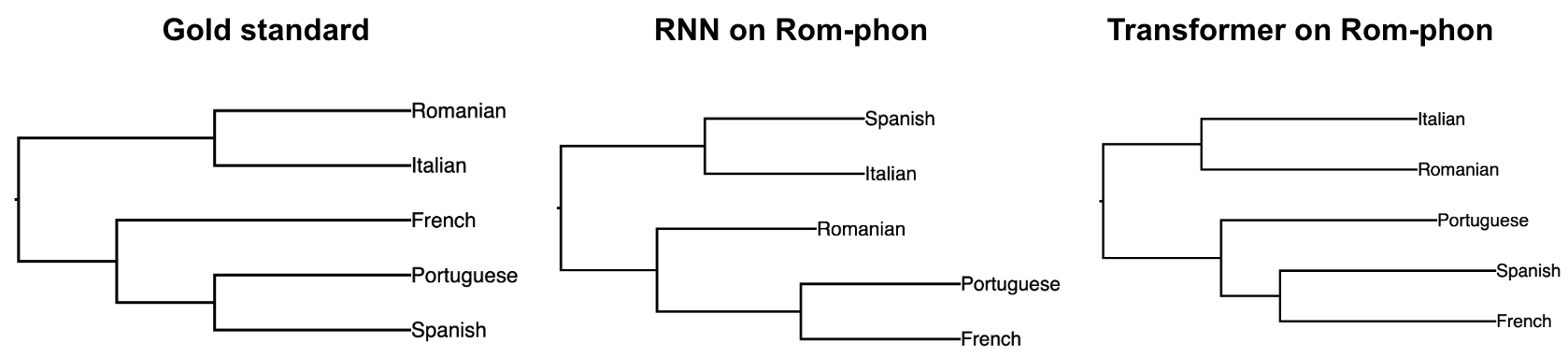}
    \caption{\label{fig:dedro-comp} A gold phylogeny of Romance (left) compared with those derived by probing the RNN model (middle) and the Transformer model (right) on Rom-phon.
    }
\end{figure*}

\section{Results and Discussion}


\subsection{Evaluation criteria}\label{sec:eval}

We evaluate the predicted protoforms using edit distance \citep{edit-dist}, normalized edit distance (edit distance normalized by the length of the target) and accuracy (the percentage of protoforms that are reconstructed without any mistakes). Like \citet{wikihan}, we also use feature error rate calculated using articulatory feature vectors from PanPhon \citep{panphon} because it reflects the phonetic similarity between the prediction and the gold protoform. For datasets with phonetic transcriptions (Romance-phonetic and Chinese), we use phoneme edit distance and normalized phoneme edit distance. As \citet{list-bcfs} suggests, we use B-Cubed F Scores \citep{bcfs} to capture the structural similarity between the gold and predicted protoforms (0: structurally dissimilar, 1: similar). 
With the exception of character and phoneme edit distance, the metrics enable fair comparison across different language families, which will differ in the average word length. 

\subsection{Results}

\autoref{tab:results} shows that our model consistently has the best performance on all datasets with regards to most metrics. The results were averaged across 10 runs.
Out of all datasets, our model performs best on the Rom-orth dataset, where we achieve a 6.55\% decrease in phoneme edit distance and a 1.41p.p improvement in accuracy relative to the RNN baseline. 
We observe the most dramatic performance difference with the RNN baseline on the Sinitic dataset: a 8.45\% decrease in phoneme edit distance and a 4.03p.p increase in accuracy. 
For reproducibility, results on the publicly available portion of the Rom-phon and Rom-orth datasets are provided in \autoref{tab:results-meloni-5000} in the Appendix.

\subsection{Analysis}

We observe that the BCFS is relatively high for the Romance non-neural baselines compared to those of the Chinese ones.
This suggests that the sound changes in the Romance datasets are more regular than that of Chinese, which corroborates \citet{list2014using}'s results that more than half of the Chinese characters in their dataset could not be explained by a tree model. 



 We examine the errors made by the Transformer model on the Rom-phon datasest. Substitutions constitute around 61\% of the errors made by the Transformer; deletions, 21\%, and insertions, 18\%. 
 The highest number of substitution errors occur between [i, \textipa{I}], [e, \textipa{E}], [o, \textipa{O}] and [u, \textipa{U}]---vowel pairs that contrast only in tenseness. This is consistent with the analysis of \citet{meloni}, where substitutions between tense-lax vowel pairs take up the largest portion of errors.
 
 We observe that other common substitution errors also happen between phonemes that share major phonetic features. This demonstrates that although no explicit phonetic information is fed directly into the model, the model makes mistakes motivated by phonetic similarity, like \citet{meloni}. 
 
We do not observe notable differences in the error statistics between the Transformer and the RNN.


\subsection{Language relatedness}

Inspired by \citet{fourrier-thesis}, we probe our model for diachronic information on how genetically related each Romance language is to each other. 
We create a distance matrix between every pair of languages in a dataset by taking the cosine similarity between a pair's language embeddings. 
We then use sklearn \citep{sklearn}'s implementation of the Ward variance minimization algorithm \citep{ward} to perform hierarchical clustering on the distance matrix. 
We take a consensus of the dendrograms from 10 different runs using the \texttt{consense} program from PHYLIP \cite{phylip}.

As we see in \autoref{fig:dedro-comp}, 
the Transformer captures more of the phylogenetic relationships among the languages correctly for the Rom-phon dataset.
Indeed, the Generalized Quartet Distance (GQD) \cite{gqd,pompei-gqd,rama2018automatic} between the gold and predicted tree, calculated using quartetDist from the \texttt{tqDist} library \cite{tqdist}, is 0.4 for the Transformer but 0.8 for the RNN. 
See Figure~\ref{fig:dedro-comp-rom-orto} in the Appendix for the results of the orthographic dataset. 




Since the Romance dataset only includes 5 daughter languages, our results are insufficient to corroborate or contradict \citet{cathcart-wandl-2020-search}'s findings: the more accurate the protoforms, the less accurate the phylogeny will be. 
It is not clear if the model's language embeddings are learning information that reflects shared innovations (sound changes that if shared among a set of daughter languages, would be acceptable justification for grouping them)---the only acceptable criterion for phylogenetic inference in historical linguistics \cite{campbell}---or if the model is learning superficial phonetic similarity.

\section{Conclusion}

By showing that Transformers can outperform previous architectures in protoform reconstruction despite the inherent data scarcity of the task, our work motivates future research in this area to take full advantage of the recent advancements in the Transformer space. 

Accurate supervised reconstruction can help predict protoforms for cognate sets where linguists have not reconstructed one yet. 
Future work could reconstruct proto-languages whose linguist reconstructions are not available, by transferring knowledge learned from languages with already reconstructed protoforms. 
Furthermore, future work can leverage the abundance of work in unsupervised NMT to adapt our Transformer model for the unsupervised setting, a more realistic scenario for the historical linguist.

\section*{Limitations}
One limitation of our work is that the RNN \citep{meloni} actually outperforms our Transformer on the Chinese dataset in \citet{wikihan}. In addition, as with other neural approaches, our model requires significant amounts of data, which is often not available to historical linguists researching less well-studied language families based on field reports. Romance and Chinese have relatively many cognate sets because the protoforms are documented\footnote{In the case of Chinese, only equivalence classes of pronunciations and not exact pronunciations are recorded.}, but a low resource setup with 200 cognate sets would not fare well on our data-hungrier Transformer model. 
Furthermore, concatenating the entire cognate set may not work on language families with hundreds of languages such as Oceanic because the input sequence would be too long compared to the output protoform sequence. 

Finally, we obtain our Chinese gold protoforms from \citet{baxter}'s Middle Chinese reconstruction, which was actually a transcription of the \textit{Qieyun}, a rhyme dictionary. 
\citet{norman-coblin} disagree with relying on such a philological source and prefer comparative reconstructions that begin from daughter data. However, there is no available comparative reconstruction of Middle Chinese with protoforms corresponding to thousands of characters to use as a gold standard. Be that as it may, it seems clear that Middle Chinese as recorded in the \textit{Qieyun} is not identical to the most recent ancestor of the Chinese languages. Its preface concedes that it is a compromise between Tang Dynasty dialects. The situation with Romance is, in some ways, comparable. Classical Latin---the variety on which we train---is not the direct ancestor of modern Romance languages. Instead, they are descended from Vulgar Latin or Proto-Romance, which is not well-attested and is primarily through graffiti and other informal inscriptions. 
Proto-Romance reconstructions are also not exhaustive. 
As a result, it is difficult to find a dataset like \citet{meloni} with thousands of such ancestor forms. 
We are also limited to the faithfulness of espeak-ng's Latin G2P, from which \citet{meloni} obtain their phonetic Romance dataset.


For most language families, protoforms are not attested. In fact, as the term is often used, protoform refers to a form that is inferred only through linguists' comparative method. We adopt the other usage for simplicity. In practice, our approach would require reconstructions made by a linguist to serve as training labels for cognate sets.







\section*{Acknowledgements}

We would like to thank Liang (Leon) Lu for finding a bug in our implementation, Ying Chen for writing the code for the baselines, and Brendon Boldt and Graham Neubig for providing useful feedback for the first iteration of our paper. 

\bibliography{acl2020}

\newpage
\appendix

\section{Training}
We split 70\%, 10\%, and 20\% of our dataset into train, validation, and test sets, respectively. We conduct hyperparameter searches using WandB \citep{wandb} and use early stopping, picking the epoch with lowest edit distance on validation data. All experiments are performed on a Ubuntu server with 4 GPUs and 20 CPUs. For both the RNN and the Transformer, \citet{meloni}'s dataset takes less than 7 GPU hours to run, while \citet{Hou2004} takes less than 1 GPU hour. For the full Romance orthographic dataset, the RNN model has 304,151 parameters, while the Transformer has 812,986 parameters. For the Romance phonetic dataset, the RNN has around 661,803 parameters, and the Transformer has around 818,640 parameters. For the Chinese dataset, the RNN has around 216,819 parameters, while the Transformer has around 2,010,967 parameters.



\section{Hyper-parameters}

Refer to \autoref{tab:hyperparam-transformer} and \autoref{tab:hyperparam-rnn} for the best hyperparameters we found during hyperparameter search via WandB.

\begin{table}[htb]
    \centering
      \SetTblrInner{rowsep=0pt}
      \begin{tblr}{lr@{}lr@{}lr@{}}
        \toprule
        & & Romance (phon \& orth) & & Sinitic\\
        \midrule
        learning rate & & 0.00013 & & 0.0007487\\
        num\_encoder\_layers & & 3 & & 2\\
        num\_decoder\_layers & & 3 & & 5\\
        embedding size && 128 && 128\\
        n\_head && 8 && 8\\
        dim\_feedforward && 128 && 647\\
        dropout && 0.202 && 0.1708861\\
        training epochs && 200 && 200\\
        warmup epochs && 50 && 32\\
        weight decay && 0 && 0.0000001\\
        batch size && 1 && 32\\
        
        \bottomrule
      \end{tblr}
      \caption{Hyper-parameters used in training the Transformer}
      \label{tab:hyperparam-transformer}
    \end{table}

\begin{table}[htb]
      \centering
      \SetTblrInner{rowsep=0pt}
      \begin{tblr}{lr@{}lr@{}lr@{}lr@{}}
        \toprule
        & & Romance-phon & & Romance-orth & & Sinitic\\
        \midrule
        learning rate & & 0.00055739 & & 0.000964 && 0.000864\\
        num\_encoder\_layers & & 1 & & 1 && 1\\
        num\_decoder\_layers & & 1 & & 1 && 1\\
        embedding size && 107 && 51 && 78\\
        hidden size && 185 && 130 && 73\\
        dim\_feedforward && 147 && 111 && 136\\
        dropout && 0.1808 && 0.323794 && 0.321639\\
        training epochs && 181 && 193 && 237\\
        warmup epochs && 15 && 15 && 15\\
        batch size && 8 && 8 && 4\\
        
        \bottomrule
      \end{tblr}
      \caption{Hyper-parameters used in training the RNN
      }
      \label{tab:hyperparam-rnn}
    \end{table}




\section{Supplementary Results}

In order to compare our model to earlier work, we used the Rom-phon and Rom-orth datasets from \citet{meloni}. However, this set includes a subset from \citet{ciobanu-dinu-2018-ab} which is not freely redistributable. So that our results can be reproduced, we also computed them on the publicly available subset of \citet{meloni}'s  dataset, which is presented in \autoref{tab:results-meloni-5000}.






\begin{table*}
\centering
\begin{tblr}{
  colspec={lXllllll}
}
\toprule
\textbf{Dataset} & \textbf{Model} & \textbf{PED $\downarrow$} & \textbf{NPED $\downarrow$} & \textbf{Acc \% $\uparrow$} & \textbf{FER $\downarrow$} & \textbf{BCFS $\uparrow$} \\
\midrule
    Rom-phon & Random daughter \cite{wikihan} & 7.1880 & 0.8201 & 0\% & 1.1396 & 0.3406 \\
    & CorPaR + Ini \cite{list-svm} & 2.0885 & 0.2491 & 14.29\% & 0.0874 & 0.6799 \\
    & SVM + PosStrIni \cite{list-svm} & 1.9005 & 0.2276 & 17.05\% & 0.0883 & 0.7039   \\
    & RNN \citep{meloni} & 1.4581 & 0.1815 & 36.68 \% & 0.0592 & 0.7435 \\
    & Transformer (present work) & 1.2516 & 0.1573 & 41.38\% & 0.0550 & 0.7790 \\
    \midrule
    Rom-orth & Random daughter \cite{wikihan} & 6.3272 & 0.6542 & 0.55\% & $-$ & 0.4023  \\
    & CorPaR + PosStrIni \cite{list-svm} & 1.8313 & 0.2001 & 18.89\% & $-$ & 0.7227 \\
    & SVM + PosStr \cite{list-svm} & 1.6995 & 0.1867 & 21.66\% & $-$ & 0.7454 \\
    & RNN \citep{meloni} & 1.3189 & 0.1505 & 38.89\% & $-$ & 0.7742\\
    & Transformer (present work) & 1.1622 & 0.1343 & 45.53\% & $-$ & 0.7989 \\
\bottomrule
\end{tblr}
\caption[Results]{Evaluation of models and baselines with various metrics on \emph{\citet{meloni}'s Romance} datasets, where all entries from \citet{ciobanu-dataset} are removed, for 1 run (using the hyperparameters of the best run on the full dataset)
}
\label{tab:results-meloni-5000}
\end{table*}



\begin{figure*}
    \centering
    \includegraphics[width=\linewidth]{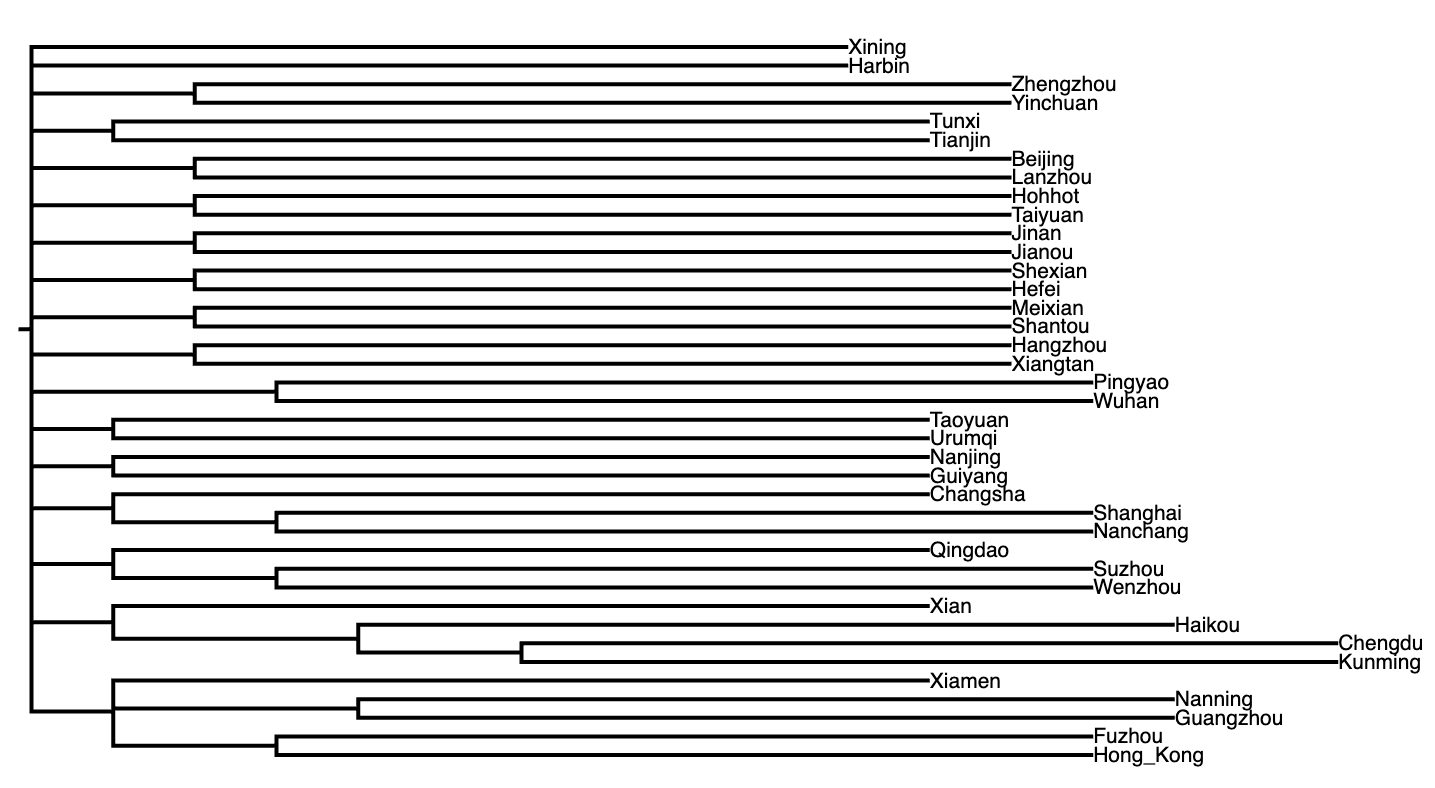}
    \caption{Consensus tree of the dendrograms from the 10 runs of the Transformer for the Chinese dataset. 
    }
    \label{fig:transformer_chinese_dendrogram}
\end{figure*}

\begin{figure*}
    \centering
    \includegraphics[width=\linewidth]{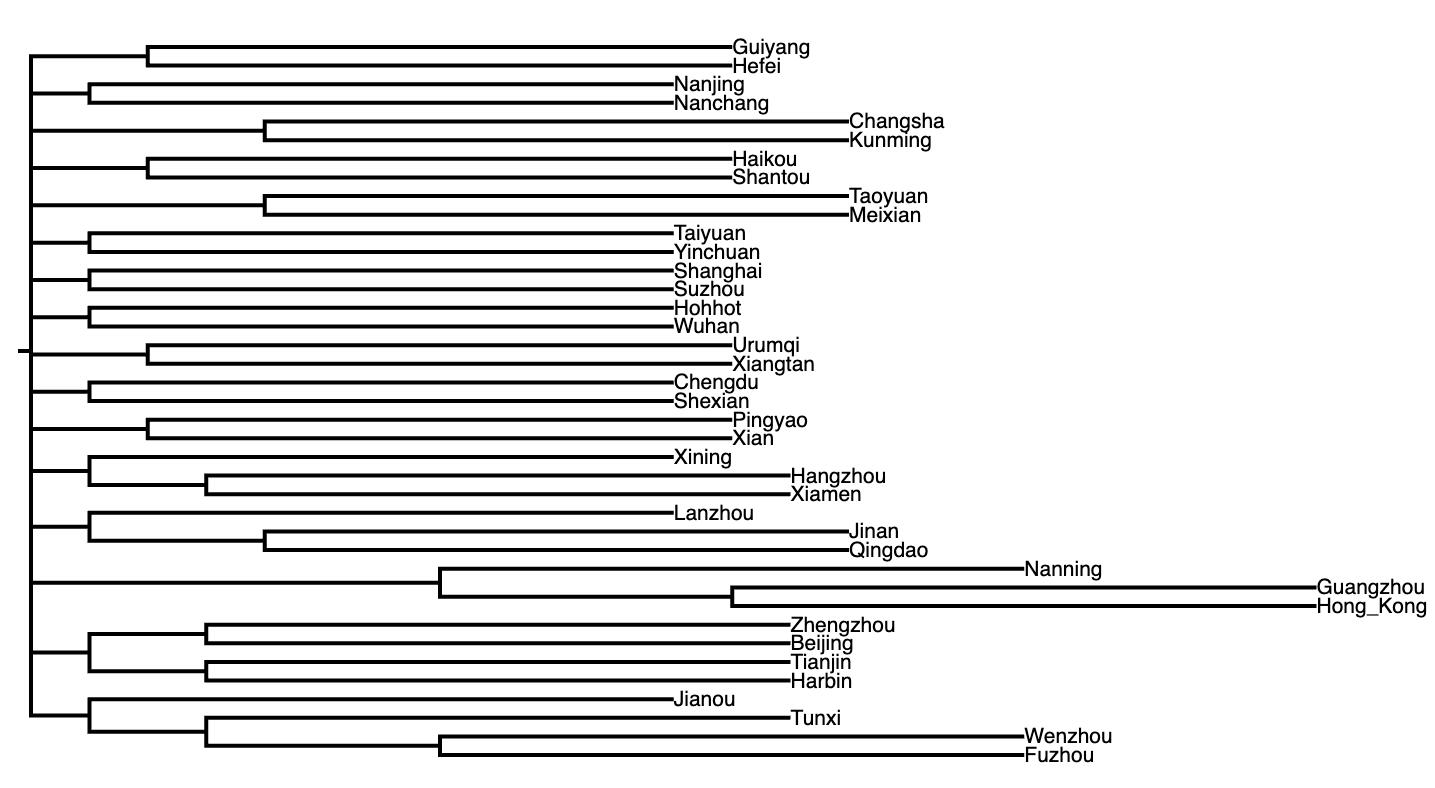}
    \caption{Consensus tree of the dendrograms from the 10 runs of the RNN for the Chinese dataset
    }
    \label{fig:rnn_chinese_dendrogram}
\end{figure*}

\begin{figure*}[h!]
    \centering
    \includegraphics[width=0.99\linewidth]{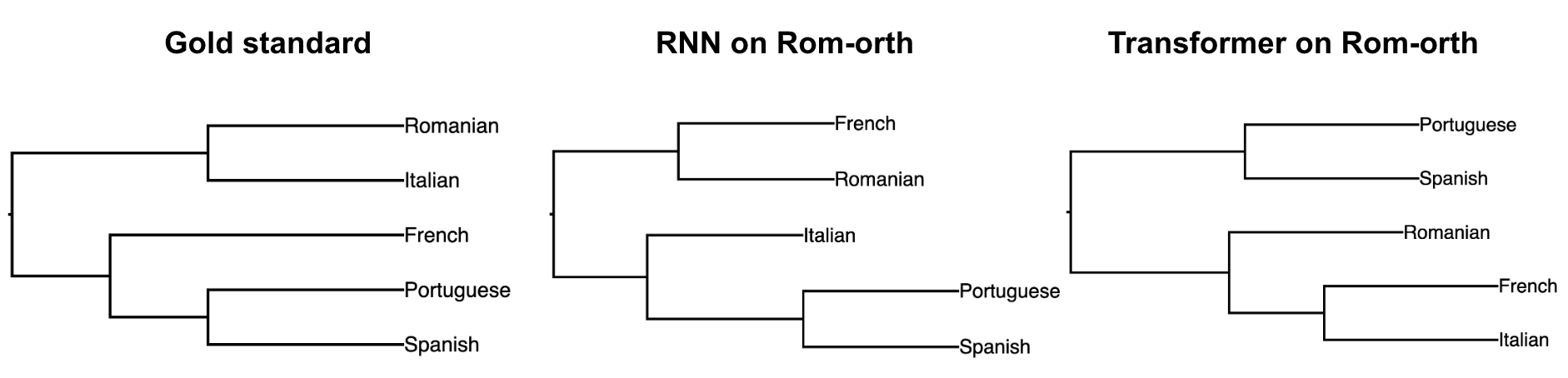}
    \caption{\label{fig:dedro-comp-rom-orto} A gold phylogeny of Romance (left) compared with those derived by probing the RNN model (middle) and the Transformer model (right) on Rom-orth. 
    GQD is 0.4 for both models.
    }
\end{figure*}

Phylogenetic trees for Chinese were also extracted from the RNN and Transformer models. These are shown in Figures~\ref{fig:transformer_chinese_dendrogram} and \ref{fig:rnn_chinese_dendrogram}. 
We also plot the dendrograms derived from the Rom-orto dataset in \autoref{fig:dedro-comp-rom-orto}. 

\end{document}